\documentclass[runningheads]{llncs}
\usepackage[T1]{fontenc}
\usepackage{graphicx}
\usepackage{tabularx}
\usepackage{subcaption}
\usepackage{float}
\usepackage{breakcites}

\begin{document}
\title{ToM-LM: Delegating Theory of Mind Reasoning to External Symbolic Executors in Large Language Models}
\titlerunning{ToM-LM: Delegating ToM to External Symbolic Executors in LLMs}
\author{Weizhi Tang\orcidID{0009-0006-7444-292X} \and Vaishak Belle\orcidID{0000-0001-5573-8465}}
\institute{
University of Edinburgh, Edinburgh, UK\\
\email{\{Weizhi.Tang, vbelle\}@ed.ac.uk}
}
\maketitle
\begin{abstract}
Theory of Mind (ToM) refers to the ability of individuals to attribute mental states to others. While Large Language Models (LLMs) have shown some promise with ToM ability, they still struggle with complex ToM reasoning. Our approach leverages an external symbolic executor, specifically the SMCDEL model checker, and fine-tuning to improve the ToM reasoning ability of LLMs. In our approach, an LLM is first fine-tuned through pairs of natural language and symbolic formulation representation of ToM problems and is then instructed to generate the symbolic formulation with a one-shot in-context example. The generated symbolic formulation is then executed by the SMCDEL model checker to perform transparent and verifiable ToM reasoning and give the final result. We demonstrate that our approach, ToM-LM, shows a significant improvement over all the constructed baselines. Our study proposes a novel view about externalizing a particular component of ToM reasoning, mainly reasoning about beliefs, and suggests generalizing it to other aspects of ToM reasoning.

\end{abstract}
\section{Introduction}
Theory of Mind (ToM) refers to an individual's ability to attribute mental states to others~\cite{Premack_Woodruff_1978,Baron-Cohen_Leslie_Frith_1985,Quesque_Rossetti_2020}. This ability is essential for individuals to understand and reason about others' beliefs, intentions, emotions, obligations, and so on~\cite{Premack_Woodruff_1978,Frith_Frith_2006,Quesque_Rossetti_2020}. For Large Language Models (LLMs), in most scenarios, they are required to interact with users or other agents and to understand their needs to give effective and appropriate responses. Therefore, discovering and enhancing their ToM reasoning ability is necessary and critical. Furthermore, it is worth noting that reasoning about knowledge has been a concern in symbolic philosophy and logic for many decades~\cite{fagin2004reasoning} and certainly in computer science to model security and game theory~\cite{Abraham_Alvisi_Halpern_2011,Halpern_2001,Halpern_Pass_Raman_2009}. To a large extent, it is also the case that most of the machine learning models today have not really supported epistemic modeling and learning. Therefore, the idea that capabilities classically dominated in symbolic computing could be used in the context of LLMs is interesting and presents intriguing possibilities.

Previous studies have demonstrated the emergence of ToM reasoning ability in LLMs~\cite{Kosinski_2023,Moghaddam_Honey,Jamali_Williams_Cai,Li_Chong_Stepputtis_Campbell_Hughes_Lewis_Sycara_2023}. Nevertheless, most of these studies adopt the ToM evaluation problems originally designed for assessing human ToM reasoning ability, which could be widely exposed to the internet, leading to possible data leakage to the training processes of LLMs~\cite{Ma_Sansom_Peng_Chai_2023,Sileo_Lernould_2023}. In addition, because of the simplicity of these ToM problems, in order to study whether LLMs can handle more complex ToM reasoning ability, new benchmarks are needed~\cite{Ma_Sansom_Peng_Chai_2023,Sileo_Lernould_2023}. Recent work has developed a new benchmark called MindGames~\cite{Sileo_Lernould_2023} aiming to give a more complex ToM benchmark specifically designed for evaluating the ToM reasoning ability of LLMs. This benchmark also indicates that despite increases in model size, LLMs still struggle with complex ToM reasoning tasks~\cite{Sileo_Lernould_2023}.

To reiterate, although ToM is widely recognized and discussed as an inherent ability in LLMs, we propose a new perspective, externalizing and delegating ToM reasoning process to external symbolic executors. Inspired by Logic-LM, which enhances the logical reasoning abilities of LLMs by leveraging external executors that provide explainable and faithful reasoning processes~\cite{Pan_Albalak_Wang_Wang_2023}, we adopt a similar approach to improve the ToM reasoning ability of LLMs. Specifically, we utilize an external symbolic executor, the SMCDEL model checker~\cite{Van_Benthem_Van_2018} designed for handling dynamic epistemic logic problems, to tackle complex ToM reasoning tasks delegated by LLMs, thereby aiming to improve the ToM reasoning ability of LLMs while making the ToM reasoning process transparent and verifiable.

\begin{figure}[!ht]
\includegraphics[width=0.95\textwidth]{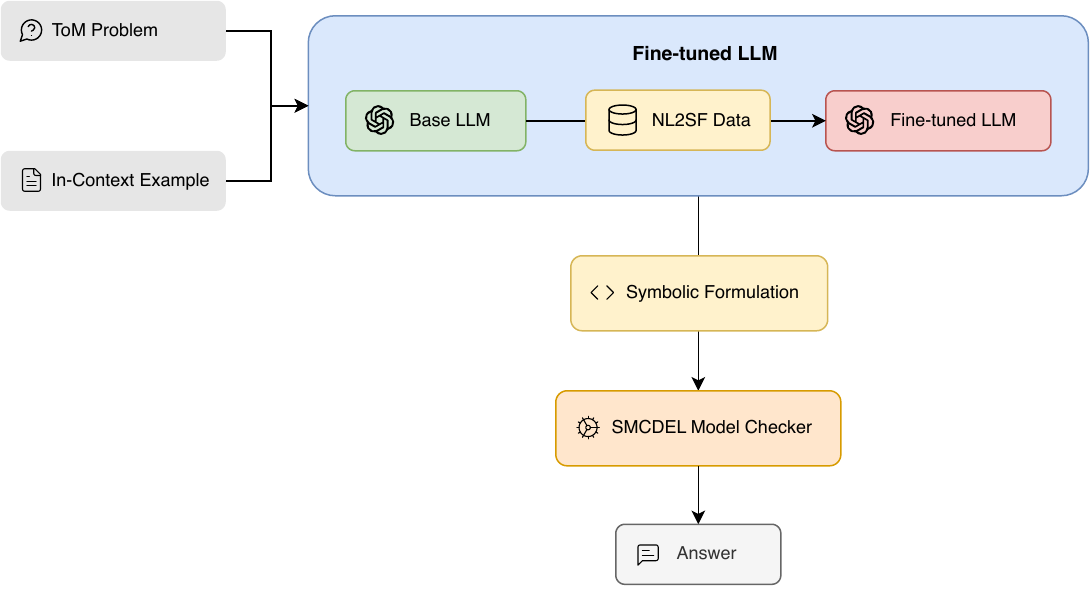}
\centering
\caption{Overview of ToM-LM, which consists of four principle stages: (1) \textit{Model Fine-tuning} in which a given base model undergoes the fine-tuning process, (2) \textit{Problem Prompting} in which a ToM problem and a one-shot example are constructed and given to the fine-tuned model, (3) \textit{Problem Formalization} in which the model generates the corresponding symbolic formulation, and (4) \textit{External Deterministic ToM Reasoning} in which the generated symbolic formulation is executed by SMCDEL model checker to give out the final result.}  \label{fig:framework}
\end{figure}

In our approach, an LLM is first given a ToM problem, and then it is asked to convert the problem represented in the natural language to the symbolic formulation. Subsequently, the generated symbolic formulation is then executed by the SMCDEL model checker to perform the ToM reasoning in a transparent and verifiable way and give out the final answer. However, our investigations reveal that this strategy alone does not adequately enhance the ToM reasoning ability of LLMs. Therefore, we employ fine-tuning\footnote{We fine-tune the chosen LLM, \textit{gpt-3.5-turbo}, through OpenAI Fine-tuning Platform with pairs of natural language and symbolic formulation. More details are discussed later in this paper.} to further optimize this approach, aiming to substantially improve the ToM reasoning ability of LLMs. The overview of ToM-LM is shown in Figure \ref{fig:framework}.

Our study demonstrates that ToM-LM significantly improves the ToM reasoning ability of LLMs in terms of accuracy and Area under the ROC Curve (AUC), compared to all constructed baselines. The contributions of this study are outlined as follows:

\begin{enumerate}
  \item We introduce a new perspective to enhance ToM reasoning ability of LLMs, externalizing and delegating ToM reasoning process to external executors to make it transparent and explainable.
  \item Our study also demonstrates that this approach makes substantial enhancements over the constructed baselines.
\end{enumerate}

\section{Related Work}
\subsection{ToM In LLMs}
 Previous works have discussed the emergence of ToM abilities in LLMs~\cite{Kosinski_2023,Jamali_Williams_Cai,Sap_LeBras_Fried_Choi_2023,trott2023large,bubeck2023sparks}. Some works also try in various ways to improve the ToM reasoning ability of LLMs. As in~\cite{Moghaddam_Honey}, the various in-context learning methods are utilized such as \textit{Let's think step by step}~\cite{Kojima_Gu_Reid_Matsuo_Iwasawa_2023} and the chain-of-thought~\cite{Wei_Wang_Schuurmans_Bosma_Ichter_Xia_Chi_Le_Zhou_2022} to enhance the ToM reasoning ability of LLMs. Although this approach brings up an easy and flexible way to improve LLMs' ToM reasoning ability, we still hold the view that dynamic and complex ToM problems can not be simply represented as a chain-of-thought reasoning process and a more appropriate and robust way to handle them is needed. Thus, we propose an approach to use symbolic formulation to represent a general way to capture the ToM reasoning process and use external symbolic executors to perform the reasoning.
\subsection{Delegating to External Executors}
Our approach is significantly inspired by~\cite{Schick_Dwivedi-Yu_Dessì_Raileanu_Lomeli_Zettlemoyer_Cancedda_Scialom_2023} which demonstrates delegating complex functionalities to external tools and by~\cite{Pan_Albalak_Wang_Wang_2023,Olausson_Gu_Lipkin_Zhang_Solar-Lezama_Tenenbaum_Levy_2023} which leverage symbolic formulation generation of LLMs and external symbolic executors to enhance their logical reasoning abilities. We follow a similar methodology and algorithmic strategy and hypothesize that it should be also suitable for the enhancement of complex ToM reasoning.

\section{Methodology}

\subsection{ToM-LM}
The overview of ToM-LM is shown in Figure \ref{fig:framework}. It consists of four principle stages, each of which is delineated in detail as follows.

\subsubsection{Model Fine-tuning} During this stage, we fine-tune the LLM, specifically the \textit{gpt-3.5-turbo}, using our newly constructed \textit{Symbolic Formulation Generation Prompting Fine-tuning Dataset}\footnote{\textit{Symbolic Formulation Generation Prompting Fine-tuning Dataset} is explained in Section \ref{sec:Dataset}.} which consists of pairs of ToM problems represented in both natural language and symbolic formulation. The fine-tuning was being processed on the OpenAI Fine-tuning Platform with 3 epochs. This process aims to improve the performance of the generation of symbolic formulations.

\subsubsection{Problem Prompting} Subsequently, a ToM problem and a one-shot in-context example are constructed as a prompt and given to the fine-tuned LLM to instruct it to give an appropriate symbolic formulation\footnote{Details of the prompting template are described in Appendix~\ref{sec:symbolic_formulation_generation}.}. 

\subsubsection{Problem Formalization} The fine-tuned LLM is then expected to generate the symbolic formulation to appropriately represent the problem described in the natural language. The generated symbolic formulations are anticipated to be executable\footnote{An example of the symbolic formulation is given in Appendix~\ref{sec:problem_example}.}. 

\subsubsection{External Deterministic ToM Reasoning} Finally, the generated symbolic formulation is then executed by the external symbolic executor, specifically the SMCDEL model checker, to perform the ToM reasoning in a transparent and verifiable way and give out the final result.

\subsection{Dataset}\label{sec:Dataset}
We select MindGames~\cite{Sileo_Lernould_2023} as our evaluation and fine-tuning dataset. It is a dataset designed specifically for evaluating the ToM reasoning ability of the LLMs. Each item\footnote{An example is given in Appendix~\ref{sec:problem_example}.} in the dataset represents a ToM problem including a premise which describes the ToM problem situation and context, a hypothesis which describes a situation needing LLMs to judge its validity, a Boolean label which is the ground truth of the validity of the hypothesis, and a corresponding symbolic formulation to the premise and hypothesis. Each problem in MindGames can be categorized as either a true-belief or false-belief problem. LLMs are expected to handle them correctly to show their ToM reasoning ability.

We then randomly sample 200 items from the test set of MindGames dataset, as the evaluation dataset, to ensure an even distribution of labels. It means the ground truth labels in this sampled evaluation dataset are meticulously balanced to achieve an exact distribution of 50\% \textit{True} label and 50\% \textit{False} label.

Additionally, we randomly sample 150 items from the MindGames train set for fine-tuning. The dataset is preprocessed in two ways for different purposes: (1) \textit{Direct Prompting Fine-tuning Dataset}, in which each problem is formatted using the \textit{Direct Prompting Template}\footnote{Details of the prompting template are described in Appendix~\ref{sec:direct_prompting_template}.} to generate prompts prepared for the DP$_{FT}$ baseline; (2) \textit{Symbolic Formulation Generation Prompting Fine-tuning Dataset}, in which each problem is formatted using the \textit{Symbolic Formulation Generation Prompting Template}\footnote{Details of the prompting template are described in Appendix~\ref{sec:symbolic_formulation_generation}.} to generate prompts prepared for ToM-LM.

\subsection{Baselines}
We choose \textit{gpt-3.5-turbo} as the base model and form three baselines to compare with our approach. The baselines are then differentiated as follows: (1) \textit{Direct Prompting} or \textit{Symbolic Formulation Generation Prompting}, and (2) whether or not Fine-tuning is applied. Our approach, ToM-LM, can be recognized as \textit{Symbolic Formulation Generation Prompting with Fine-tuning}.

\subsubsection{DP}
It stands for \textit{Direct Prompting Without Fine-tuning}. In this setting, we directly ask the base model to answer the problem in "TRUE", "FALSE", or "I DON'T KNOW", with a one-shot in-context example provided. Answering in "TRUE" means it considers the hypothesis is valid in the given situation, "FALSE" means it thinks the hypothesis is invalid, and "I DON'T KNOW" means it has no idea about the validity of the hypothesis.

\subsubsection{SFGP}
It stands for \textit{Symbolic Formulation Generation Prompting Without Fine-tuning}. In this setting, the workflow is the same as ToM-LM but the model is not fine-tuned.

\subsubsection{DP$_{FT}$}
It stands for \textit{Direct Prompting With Fine-tuning}. In this setting, the workflow is the same as the DP baseline but the model is fine-tuned based on the \textit{Direct Prompting Fine-tuning Dataset}.

\section{Results}

The responses given by DP, SFGP, DP$_{FT}$, and ToM-LM can be categorized and summarized into three types in general: (1) \textit{TRUE}, (2) \textit{FALSE}, or (3) \textit{I DON'T KNOW}. 

For \textit{Direct Prompting} setting, \textit{TRUE} simply means the LLM answers in "TRUE" literally, \textit{FALSE} means the LLM answers in "FALSE" literally, and answering in "I DON'T KNOW" literally or other contents are considered as answering in \textit{I DON'T KNOW}. 

For \textit{Symbolic Formulation Generation Prompting} setting, \textit{TRUE} means the LLM generates an executable symbolic formulation and the model checker executes it to get \textit{TRUE} value towards the problem, \textit{FALSE} means the LLM generates an executable symbolic formulation and the model checker executes it to get \textit{FALSE} value towards the problem, and \textit{I DON'T KNOW} in this case means the LLM fails to generate an executable symbolic formulation.

\subsection{Metrics}
The evaluation metrics consist of three aspects, (1) Execution Rate, (2) Accuracy, and (3) Area under the ROC Curve (AUC). Table \ref{table:metrics} shows all the three metrics for DP, SFGP, DP$_{FT}$, and ToM-LM.

\begin{table}
\caption{The Metrics of DP, SFGP, DP$_{FT}$, and ToM-LM}\label{table:metrics}
\begin{center}
\begin{tabularx}{\textwidth}{|X|X|X|X|}
\hline
Approach &  Execution Rate(\%) & Accuracy(\%) & AUC \\
\hline
DP &  99.50 & 58.00 & 0.58\\
SFGP& 78.00 & 49.00 & 0.60\\
DP$_{FT}$ & \textbf{100} & 76.00 & 0.76\\
ToM-LM & 94.50 & \textbf{91.00} & \textbf{0.94} \\
\hline
\end{tabularx}
\end{center}
\end{table}

\subsubsection{Execution Rate} 
This metric is used to measure how successfully the LLM can generate available or executable answers. In the case of \textit{Direct Prompting} setting, it is \textit{TRUE} or \textit{FALSE} and in the case of \textit{Symbolic Formulation Generation Prompting} setting, it is executable symbolic formulation. A higher value of execution rate means higher available or executable answers generated by the LLM. But it does not mean the generated answers are correct.

The execution rates of baselines in \textit{Direct Prompting} setting are almost around 100\%, which means the LLM can generate available answers at most of time in this setting. Instead, in the case of \textit{Symbolic Formulation Generation Prompting} setting, the execution rate is 78\% without fine-tuning but can be high up to 94.5\% in ToM-LM.

\subsubsection{Accuracy} As demonstrated in Table \ref{table:metrics}, ToM-LM reaches up to 91\% accuracy which significantly surpasses all the baselines. It achieves the 33\% improvement over the DP baseline, the 42\% increase compared to the SFGP baseline, the 15\% enhancement over the DP$_{FT}$ baseline.

\subsubsection{AUC} Additionally, AUC is utilized to provide a more holistic indicator and assessment of the performance of ToM-LM and the baselines. As shown in Table \ref{table:metrics}, while the AUC for DP and SFGP are around 0.60, ToM-LM reaches up to the 0.94 AUC and outperforms all the baselines.

\subsection{Output Distributions}
\begin{figure}
\includegraphics[width=\textwidth]{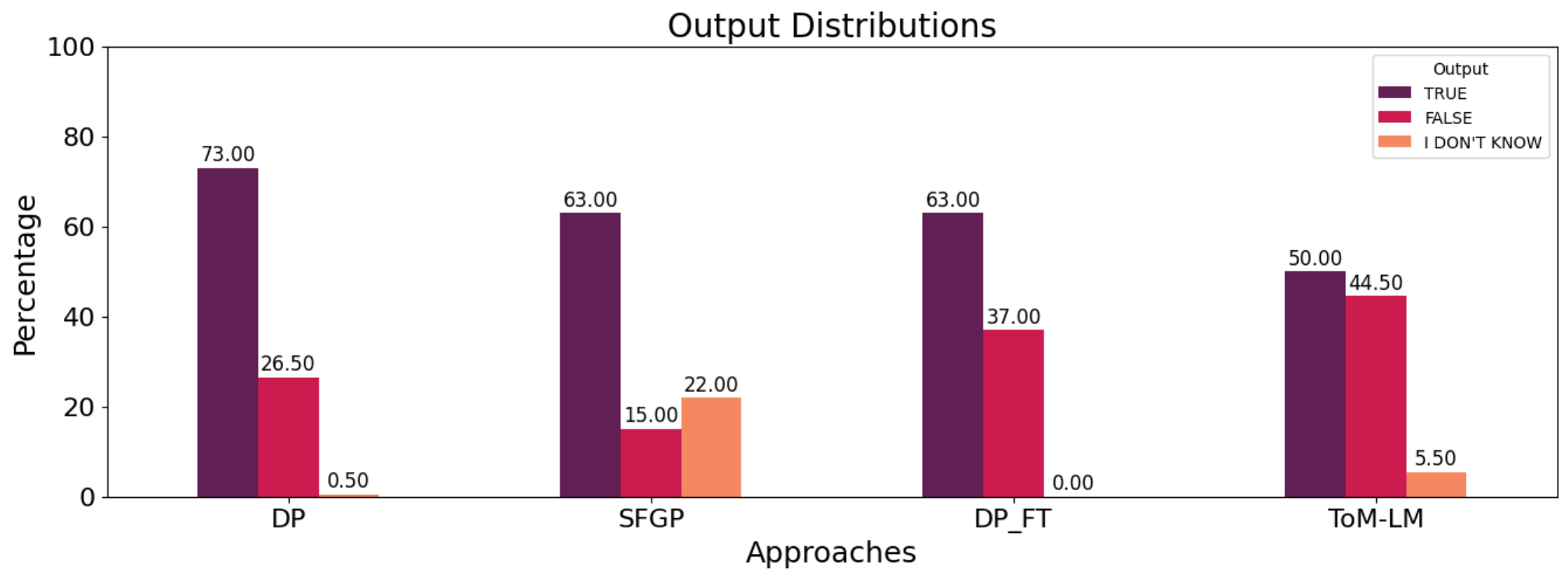}
\caption{Output Distributions of DP, SFGP, DP$_{FT}$, and ToM-LM} \label{fig:distribution}
\end{figure}

We have also illustrated the distributions of outputs in Figure \ref{fig:distribution} to provide an intuitive visualization of the results and performance. Given that the evaluation set is evenly distributed across the ground truth labels, the distributions of outputs can intuitively reflect performance differences. While the output distributions for all baseline models exhibit significant bias, the output distribution of ToM-LM is notably more balanced.

\section{Discussion}

\subsection{Results Analysis}
The results demonstrate that ToM-LM has successfully enhanced the ToM reasoning ability of LLMs, surpassing the other three baselines in terms of both accuracy, which reaches as high as 91\%, and AUC which attains 0.94.

From the results of the DP baseline, the accuracy is 58\% and AUC is 0.58, meaning although the LLM itself may possess a modest degree of ToM reasoning ability, it still struggles with it and its answers are highly dependent on random guessing. From the SFGP baseline, although AUC is shown with a marginal improvement by 0.02 and reaches 0.6, indicating that it modestly enhances the performance of the LLM against random guessing, the gain comes at the cost of reduced accuracy and a highly increased incidence of non-executable symbolic formulations.

As anticipated, the DP$_{FT}$ baseline markedly surpasses both of the DP and SFGP baseline in terms of both accuracy and AUC. When comparing it to our approach, ToM-LM shows significant improvement over it with both of accuracy and AUC. In addition, within the same fine-tuning settings, we also observe that the training loss of the DP$_{FT}$ baseline has a violent oscillation and does not converge at the end but the training loss of the ToM-LM has a very slighter oscillation and it converges at the end of fine-tuning\footnote{Loss curves for both DP$_{FT}$ and ToM-LM are given in Appendix~\ref{sec:ft_loss_curve}.}.

Furthermore, the distributions of outputs indicate a significant bias among the three baselines, with all tending to respond \textit{TRUE} to the given ToM problems. This pattern reveals their inability to accurately handle false-belief tasks, underscoring a deficiency in ToM reasoning ability. And as shown in Figure \ref{fig:distribution}, it is noteworthy that even in the SFGP baseline, the LLM fails to generate appropriate executable symbolic formulations for false-belief tasks, while predominantly generating symbolic formulations for true-belief tasks instead. However, ToM-LM significantly improves the balance of results compared to all the three baselines.

\subsection{Limitations and Future Work}
While ToM is supposed to encompass a broad range of aspects, such as emotions, goals, intentions, desires, and beyond, our study only focuses on handling false-belief and true-belief of ToM due to the expressive limitations of the current external executor. Nevertheless, this approach has the potential to be generalized to handle other ToM tasks. Future research should explore additional components of ToM using external executors and may also aim to develop a more expressive and capable external executor or language to handle a diverse array of ToM tasks.

\section{Conclusion}
In this study, we propose a novel approach, ToM-LM, to externalize the ToM reasoning process of LLMs by employing the external symbolic executor, specifically the SMCDEL model checker. It aims to improve the ToM reasoning ability of LLMs while also making the ToM reasoning process transparent and explainable. Our approach demonstrates significant improvements over the three constructed baselines. We also anticipate that this research can offer new insights about improving other aspects of ToM reasoning ability of LLMs.

\bibliographystyle{splncs04}
\bibliography{references}

\clearpage
\appendix

\section{Prompting Templates}
\label{sec:prompting_template}

\subsection{Direct Prompting Template}
\label{sec:direct_prompting_template}

\begin{figure}
\centering
\includegraphics[width=\textwidth]{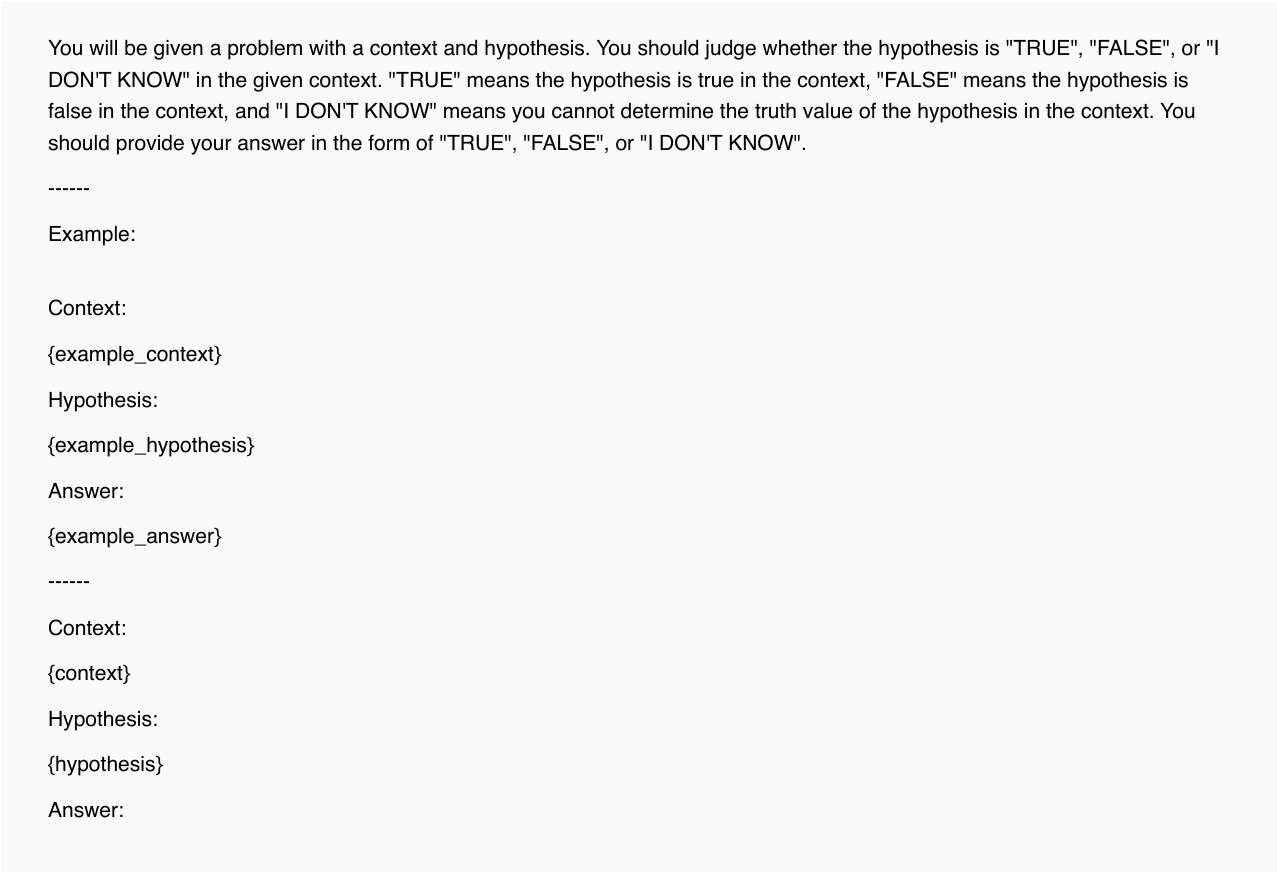}
\caption{Direct Prompting Template} \label{fig:direct_prompt}
\end{figure}

Figure \ref{fig:direct_prompt} shows the template used for prompting to directly generate the prediction of the answer. The texts within the braces serve as placeholders to be replaced with actual example or problem contents. The placeholders beginning with "example" indicate that they are for the one-shot in-context ToM example, while the placeholders "context" and "hypothesis" are for the ToM problem.

\subsection{Symbolic Formulation Generation Prompting Template}
\label{sec:symbolic_formulation_generation}

\begin{figure}
\centering
\includegraphics[width=\textwidth]{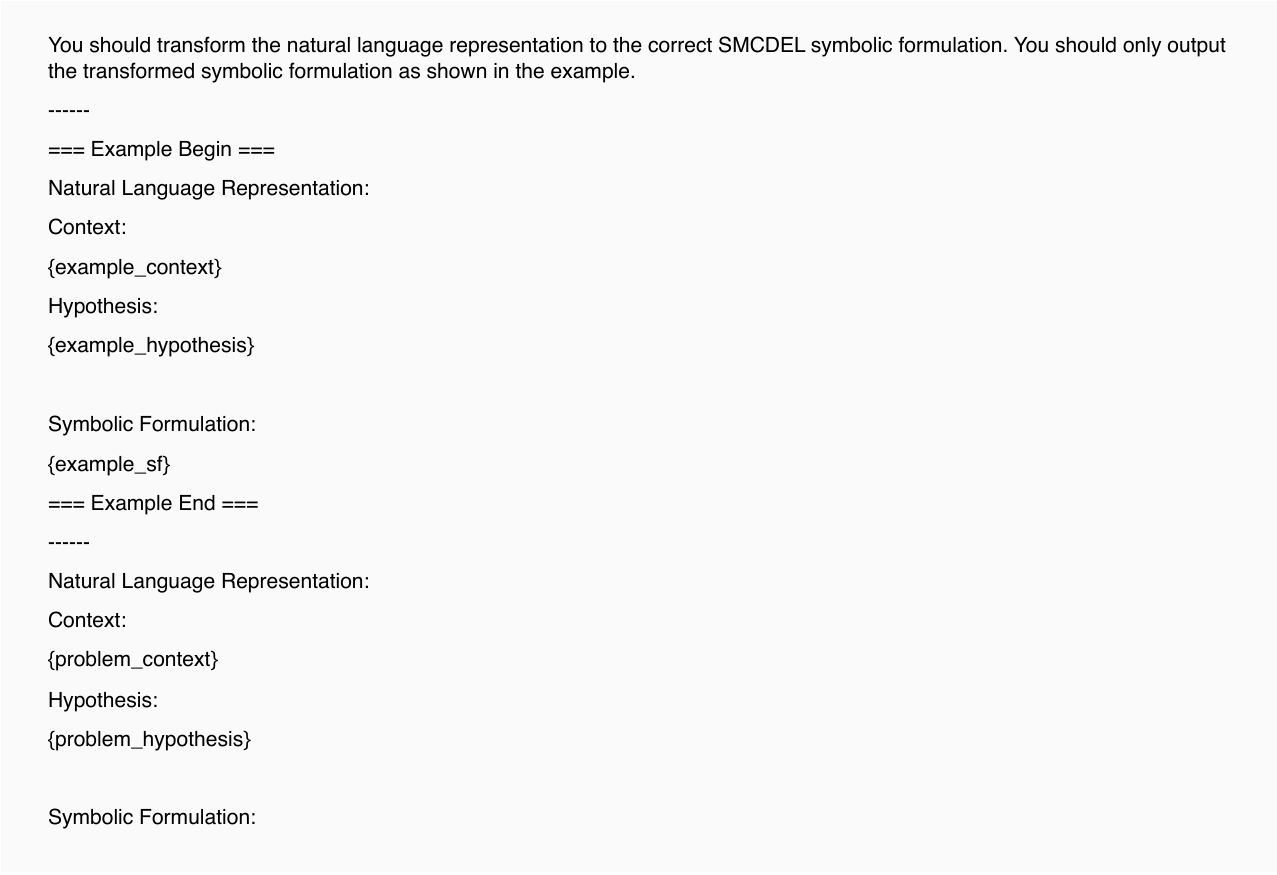}
\caption{Symbolic Formulation Generation Prompting Template}
\label{fig:sf_prompt}
\end{figure}

Figure \ref{fig:sf_prompt} demonstrates the template used for prompting to generate the prediction of the symbolic formulation. The texts within the braces serve as placeholders to be replaced with actual example or problem contents. The placeholders beginning with "example" indicate that they are for the one-shot in-context ToM example while the placeholders starting with "problem" mean they are for the ToM problem.

\section{Problem Example}
\label{sec:problem_example}

\begin{figure}
    \centering
    \includegraphics[width=\textwidth]{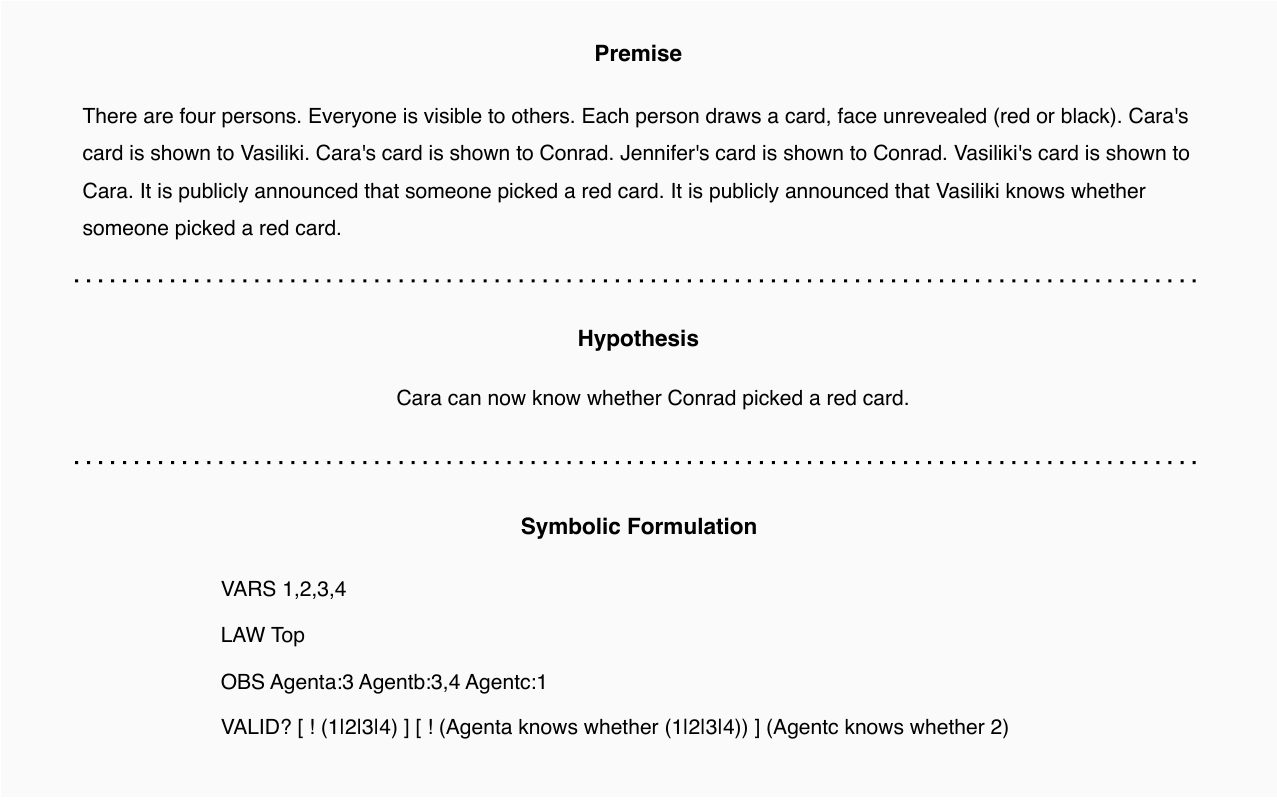}
    \caption{An example of premise, hypothesis, and symbolic formulation in MindGames.}
    \label{fig:mindgames-example}
\end{figure}

An example of a ToM problem consisting of a premise, a hypothesis, and a symbolic formulation given in MindGames~\cite{Sileo_Lernould_2023} is illustrated in Figure \ref{fig:mindgames-example}.

The premise describes a scenario involving four agents, each drawing a face unrevealed card, and someone may reveal their card to others. The hypothesis to be validated based on this given scenario is "Cara can now know whether Conrad picked a red card". The corresponding symbolic formulation representation of the premise and hypothesis is also given. To briefly explain\footnote{Refer to https://github.com/jrclogic/SMCDEL for more syntax details.}, "VARS 1,2,3,4" defines four atomic propositions which represent the status of each agent, such as that "2" represents "Conrad picked a red card". "LAW Top" indicates that all possible Boolean states of these statuses should be considered. "OBS" simply denotes which agent observes which status given in "VARS", such as that "Agentc:1" means Cara observes the status that "Vasiliki picked a red card". The "VALID?" line queries an agent's belief about a certain status, in which "!" means a public announcement to all agents and "|" is a logical \textit{OR}, such as that "!(1|2|3|4)" means "It is publicly announced that someone picked a red card". It represents and encapsulates the main part of the premise and hypothesis.

\section{Fine-tuning Loss Curves}
\label{sec:ft_loss_curve}

\subsection{Loss Curve of Direct Prompting With Fine-tuning}

\begin{figure}
    \centering
    \includegraphics[width=\textwidth]{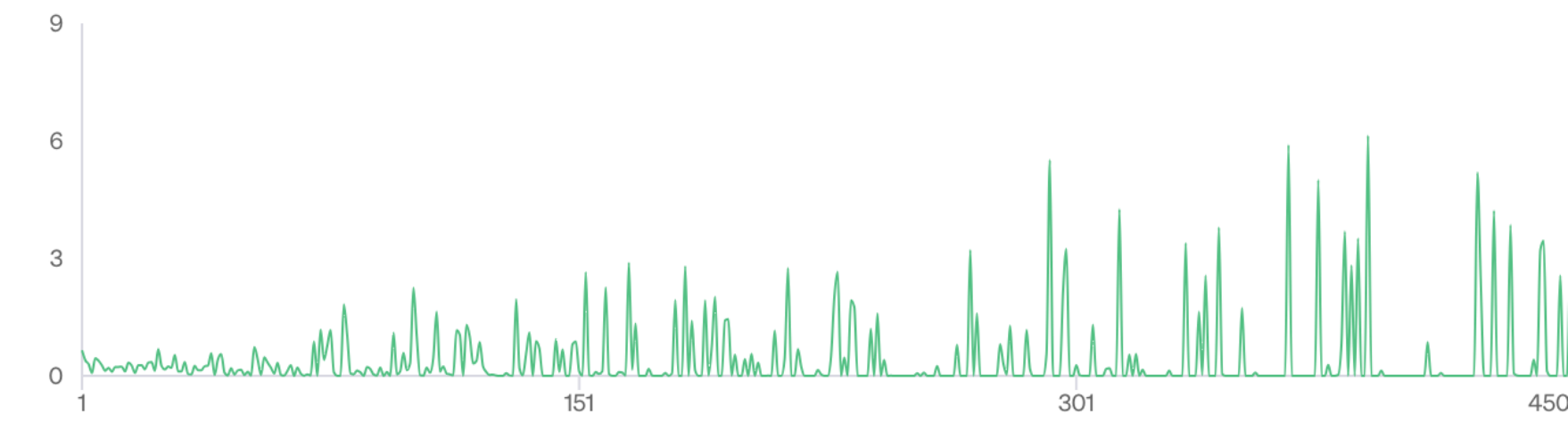}
    \caption{Loss Curve of Direct Prompting With Fine-tuning}
    \label{fig:direct_ft_loss}
\end{figure}

As the loss curve illustrated in Figure~\ref{fig:direct_ft_loss}, the training loss of the DP$_{FT}$ baseline is high with the range of loss between 0 and 9 as shown in the y-axis, while also exhibits severe oscillations and failing to converge by the end of training.

\subsection{Loss Curve of ToM-LM}

\begin{figure}
    \centering
    \includegraphics[width=\textwidth]{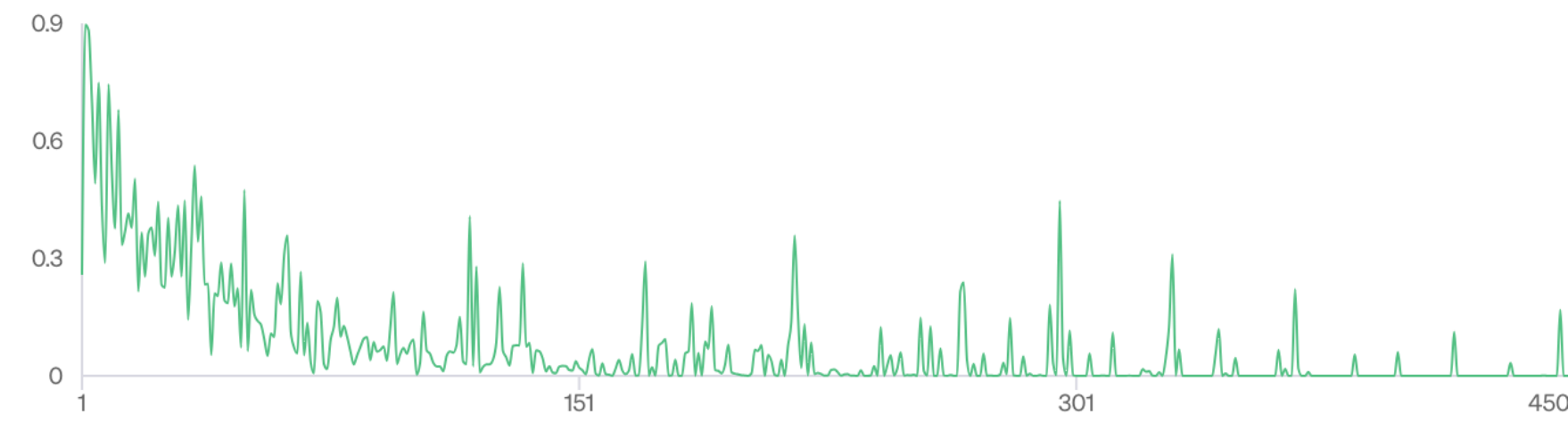}
    \caption{Loss Curve of ToM-LM}
    \label{fig:sf_ft_loss}
\end{figure}

As shown in Figure~\ref{fig:sf_ft_loss}, the training loss of the ToM-LM is low with the range of loss between 0 and 0.9 and converges at the end of training.

\end{document}